\documentclass{tlp}

\usepackage{epsfig}
\usepackage{latexsym}
\usepackage{theorem}
\usepackage{amssymb}
\usepackage{subfigure}
\usepackage{floatflt}

\newcommand{\tab}{\hspace*{1em}}
\newcommand{\txt}[1]{\mbox{\tt #1}}
\newcommand{\BEGIN}{\mbox{\sffamily\bfseries \{~}}
\newcommand{\END}{\mbox{\sffamily\bfseries \}~}}
\newcommand{\IF}{\mbox{\sffamily\bfseries if~}}

\newcommand{\ELSE}{\mbox{\sffamily\bfseries else~}}

\newcommand{\FOR}{\mbox{\sffamily\bfseries for~}}

\newcommand{\RETURN}{\mbox{\sffamily\bfseries return~}}

\newcommand{\WHILE}{\mbox{\sffamily\bfseries while~}}

\newtheorem{example}{Example} 
\newcommand{\qed}{\proofbox}

\begin{document}

\bibliographystyle{acmtrans}

\submitted{31 August 2002}
\revised{13 November 2003, 16 June 2004}
\accepted{9 August 2004}

\sloppy


\title[Intelligent search based on adaptive CHR]
{Intelligent search strategies based on adaptive Constraint Handling Rules}

\author[Armin Wolf]
{ARMIN WOLF\\
      Fraunhofer-Institut f\"ur Rechnerarchitektur and Softwaretechnik FIRST\\
      Kekul\'{e}stra\ss{}e 7, D-12489 Berlin, Germany\\
      \email{Armin.Wolf@first.fraunhofer.de~~~http://www.first.fraunhofer.de}
}

\maketitle

\begin{abstract}
  The most advanced implementation of adaptive constraint processing
  with Constraint Handling Rules (CHR) allows the application of
  intelligent search strategies to solve Constraint Satisfaction
  Problems (CSP). This presentation compares an improved version of
  conflict-directed backjumping and two variants of dynamic backtracking
  with respect to chronological backtracking on some of the AIM
  instances which are a benchmark set of random 3-SAT problems.  A CHR
  implementation of a Boolean constraint solver combined with these
  different search strategies in Java is thus being compared with a
  CHR implementation of the same Boolean constraint solver combined
  with chronological backtracking in SICStus Prolog.  This comparison
  shows that the addition of ``intelligence'' to the search process
  may reduce the number of search steps dramatically. Furthermore, the
  runtime of their Java implementations is in most cases faster than
  the implementations of chronological backtracking. More specifically,
  conflict-directed backjumping is even faster than the SICStus Prolog
  implementation of chronological backtracking, although our Java
  implementation of CHR lacks the optimisations made in the SICStus
  Prolog system.
\end{abstract}


\section{Introduction}

Constraint Handling Rules (CHR) are multiheaded, guarded rules used
to propagate new or simplify given
constraints~\cite{chrFruehwirth:95,chrFruehwirth:98}. For example,
the CHR 
\begin{center}
  {\tt leq(X,Y), leq(Y,Z) ==> leq(X,Z).}
\end{center} 
reflects the transitivity of the binary relation {\tt leq}.  Thus,
for any two constraints {\tt leq(A,B)} and {\tt leq(B,C)} an
additional constraint {\tt leq(A,C)} is derived -- implicitly given
knowledge is made explicit.  Another CHR
\begin{center}
  {\tt leq(X,Y), leq(Y,X) <=> X=Y.}
\end{center} 
reflects the symmetry of the binary relation {\tt leq}. Thus, any two
constraints {\tt leq(A,B)} and {\tt leq(B,A)} are replaced by the
syntactical equation {\tt A=B}.

A detailed formal description of the syntax, the declarative and
operational semantics of CHR is omitted in this paper because these
topics are addressed in depth in the literature, e.g.
in~\cite{chrFruehwirth:98}.

There are several CHR implementations, e.g. in
ECLiPSe~\cite{chrFruehwirth:Brisset:95}, in SICStus
Prolog~\cite{chrHolzbaur:Fruehwirth:00a} or even in
Java~\cite{javaSchmauss:99,chrWolf:01a}.  All but the
last~\cite{chrWolf:01a} only support constraint deletions implicitly
through chronological backtracking. Arbitrary sequences of constraint
additions and deletions, which are necessary for intelligent search
strategies like conflict-directed
backjumping~\cite{dcpProsser:93,dcpProsser:95} or dynamic
backtracking~\cite{dcpBaker:94,dcpGinsberg:93,dcpJussien:etal:00}, are
not supported.  Furthermore, if there is an inconsistency, the
``classical'' CHR implementations offer users no help in finding out
what causes this inconsistency.

This paper reviews the first implementation of ``adaptive'' CHR,
cf.~\cite{chrWolf:99d,chrWolf:01a}. In this context, ``adaptive''
means that constraint additions and deletions in arbitrary order are
supported, i.e. after each change of the considered constraints, the
derivations based on CHR are adapted accordingly. Thus, deletion of
the constraint {\tt leq(A,B)} or {\tt leq(B,C)} causes the derived
constraint {\tt leq(A,C)} to be deleted, too. However, this
implementation is in Java, which does not support backtracking like
Prolog systems; thus depth-first search is not intrinsic, enabling
different search strategies to be realized directly and not on top of
the underlying chronological backtracking mechanism.  Additionally,
this implementation returns an explanation for any occurring
inconsistency, thus supporting \emph{explanation-based constraint
  programming} ~\cite{dcpJussien:01}.  This allows not only user
guidance, e.g. during debugging of incorrect constraint models or in
interactive constraint solving, but also automatic constraint
relaxation as well as dynamic problem handling in reactive systems.
Moreover, explanations can be used to build new
``explanation-directed'' search algorithms.

The aim of the paper is to show that this adaptive CHR implementation
is very well suited for implementing not only depth-first search but
also intelligent search algorithms like sophisticated
conflict-directed backjumping and dynamic backtracking, while
maintaining consistency. The given implementations show that
\begin{itemize}
  \item constraint propagation ideally replaces the proposed constraint
        checks/tests in these intelligent search algorithms
  \item justifications of (derived) constraints, especially of {\tt false},
        properly act as \emph{conflict sets} in conflict-directed backjumping 
        or as \emph{elimination explanations} in dynamic backtracking
  \item the possibility of arbitrary constraint deletions directly
        supports non-chronological backtracking
  \item constraint handling (i.e. propagation) maintains (local) 
        consistency, offering early detection and good avoidance 
        of dead ends during the search
\end{itemize} 
To be more precise, the implementation of these algorithms is
specialised for Boolean constraint problems where the variables
have only two possible values: 0 and 1. However, a generalisation for
other (finite) domains is quite simple because the interaction with
the Boolean constraint solver written in CHR is opaque. We therefore 
assume that any other terminating constraint solver realized with CHR
will work as well. We cite the soundness and completeness of
CHR~\cite{chrFruehwirth:98} as well as the correctness and termination
of the adaptation of CHR derivations~\cite{chrWolf:99d} based on
explanations as evidence for this claim.

The paper is organised as follows. The next section briefly
looks at the adaptive CHR system. Section~\ref{boolSolver}
presents a CHR-based specification of a Boolean constraint solver to
solve SAT(isfiability) problems formulated as propositional logic
formulas. The compilation process of these rules into an adaptive
constraint solver is explained and the application programming
interface (API) for this solver is described. Section~\ref{aim}
introduces the AIM instances, a benchmark set of random 3-SAT problems
containing instances with exactly one solution and instances that are
inconsistent. Section~\ref{search} presents our implementations of
different search strategies, from chronological backtracking
(Section~\ref{secCBT}) to conflict-directed backjumping
(Section~\ref{secCBJ}) to dynamic backtracking (Section~\ref{secDBT}).
These implementations are built on top of the Boolean constraint solver
presented in Section~\ref{boolSolver}. These solvers are applied to all
AIM instances with 50 variables, either to solve them or to detect
their inconsistency. Section~\ref{experiments} shows and compares the
required backtracking/backjumping steps and their runtime.
Section~\ref{discussion} attempts to analyse the measured results.
Section~\ref{conclusion} concludes the paper.

\section{The Adaptive CHR System}

\begin{floatingfigure}{0.35\textwidth}
\hspace*{-0,05\textwidth}
\epsfig{file=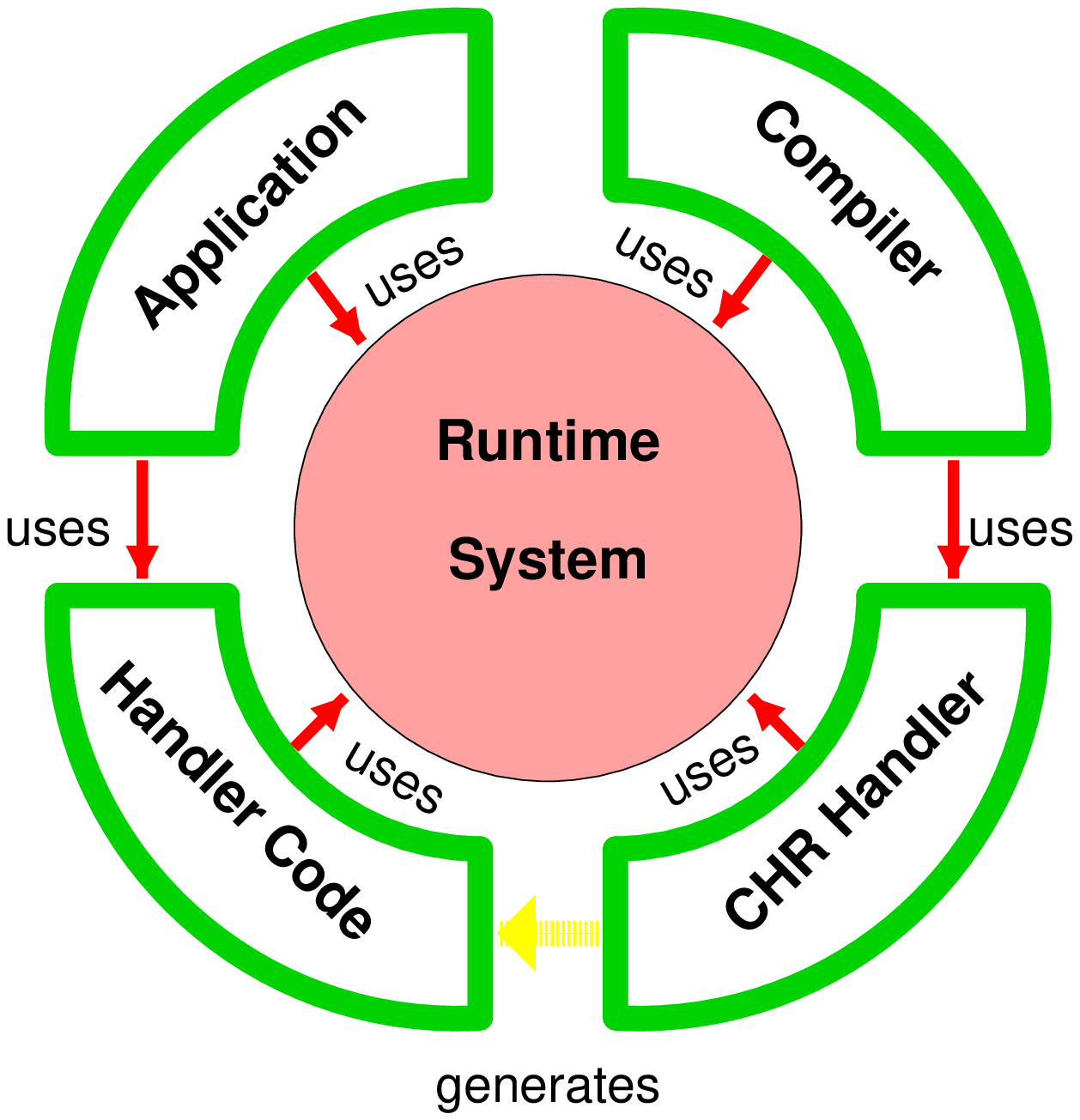, width=0.35\textwidth}
\caption{The architecture of the adaptive CHR system}\label{system}
\end{floatingfigure}
Initially, the adaptive CHR system consists of a runtime system and a
compiler. They contain the data structures that are required to
generate rule-based adaptive constraint solvers and to implement Java
programs that apply these solvers to dynamic CSP.  The definition of a
rule-based constraint solver is quite simple: the CHR that define the
solver for a specific domain are coded by the user in a so-called CHR
handler. Here, a CHR handler consists of Java objects representing CHR
which are compiled to Java programs by the use of the compiler. 
Compiling and running a CHR
handler generates a Java package containing Java code that implements
the defined solver and its interface: the addition or deletion of
user-defined constraints or syntactical equations, a consistency test
and the explanation of inconsistencies. These methods allow dynamic
constraint solving as well as \emph{explanation-based constraint
  programming}~\cite{dcpJussien:01} in any application:
\begin{itemize}
  \item Constraints may be added and deleted in arbitrary order. 
  \item Constraint handling, i.e. propagation, is performed accordingly.  
  \item Whenever an inconsistency is detected, the explanation identifies 
        a subset of constraints causing this inconsistency.
\end{itemize}

\noindent
A user application interacts with the CHR package provided by the user
in the CHR handler and the runtime system.  Figure~\ref{system} shows
the components and their interactions.

During compilation for each handler, a constraint system class is
generated retaining the name of the handler. Furthermore, for each
head constraint of a CHR, a method of this class is generated
retaining the name and arity of the CHR to add user-defined
constraints to the constraint store. In addition to these handler-specific
methods each constraint system class has common methods to
justify the assignment of an integer to a variable, i.e. to add a
syntactical equation justified by an integer to the constraint store;
to delete all constraints with a specific justification, i.e. in a set
of integers; to test the consistency of the currently valid syntactical
equations; and to get an explanation, i.e. a set of justifications
(integers) that is responsible for a detected inconsistency:
\begin{itemize}
  \item {\tt  void equal(Variable var, int i, IntegerSet set)}
  \item {\tt  void delete(IntegerSet set)}
  \item {\tt  boolean isConsistent()}
  \item {\tt  IntegerSet getExplanation()}
\end{itemize}
The class {\tt Variable} implements logical variables that may be
bound to logical terms (objects of the class {\tt Term}), which are
either numbers, logical variables or function terms. To simplify
matters, integer sets (objects of the class {\tt IntegerSet}) and
operations on it are represented by the use of the usual mathematical
set notation, e.g.  the set consisting of the integers 2, 3, and 5 is
represented by $\{2,3,5\}$ and the union of two sets $A$ and $B$ is
represented by $A \cup B$.

\begin{example}
  Let a CHR handler called {\tt trans} consist of the CHR 
  \begin{center}
    {\tt leq(X,Y), leq(Y,X) <=> X=Y.}
  \end{center}
  specifying the user-defined constraint {\tt leq}. Furthermore, let
  the constraints {\tt leq(0,A)} and {\tt leq(B,1)} with empty
  justifications be already added to the constraint store of the
  constraint system {\tt cs}, i.e. be an object of the class {\tt
    trans}.  Assuming that {\tt A} and {\tt B} are constraint
  variables (objects of the class {\tt Variable}), {\tt cs.equal(A, 2,
    $\{1\}$)} adds the equation {\tt A=2} justified by the
  set\footnote{For a unitised handling, integral identifiers are coded
    in singleton integer sets.}~$\{1\}$ to the constraint store of
  {\tt cs}.  Thus, the value {\tt 2} is assigned to the variable {\tt
    A} and the store contains {\tt A=2}, {\tt leq(0,2)}, both
  justified by $\{1\}$, and {\tt leq(B,1)} justified by the empty set.
  Then, the call {\tt cs.isConsistent()} returns true.  Further
  addition of the equation {\tt B=2} with the justification $\{3\}$ is
  realized by calling {\tt cs.equal(B, 2, $\{3\}$)}.  The resulting
  constraint store now contains {\tt A=2}, {\tt leq(0,2)}, both
  justified by $\{1\}$, and {\tt B=2} and {\tt leq(2,1)}, both
  justified by $\{3\}$.  This triggers the CHR, which replaces {\tt
    leq(0,2)} and {\tt leq(2,1)} by the equation {\tt 0=1}, i.e. an
  inconsistency justified by $\{1,3\}$.  Thus, the call {\tt
    cs.isConsistent()} returns false and {\tt cs.getExplanation()}
  returns $\{1,3\}$.  The detected inconsistency is eliminated by
  calling {\tt cs.delete($\{1\})$}.  Afterwards, the constraint store
  contains {\tt leq(A,2)} with the empty justification, and {\tt B=2}
  and {\tt leq(2,1)}, both justified by $\{3\}$.  \qed
\end{example}

The next section contains a more relevant, practical example,
illustrating how constraint solvers, i.e. CHR handlers, are defined
and integrated into an application.

\section{A Rule-based Boolean Constraint Solver}\label{boolSolver}

The ECLiPSe and SICStus Prolog distributions of CHR, or even WebCHR at
{\tt http://www.pms.informatik.uni-muenchen.de/\~{}webchr/}, come with
a simple but important constraint solver for Boolean constraints.
This solver is essential for problems that are formulated as SAT
problems, i.e. satisfiability problems of propositional logic
formulas.  The provided Boolean constraint solver supports the usual
unary and binary operations on propositional variables: negation,
conjunction, disjunction (non-exclusive and exclusive) as well as
implication.  If we confine ourselves -- without any loss of
expressiveness -- to problems in conjunctive normal form, only
negation and disjunction have to be supported by a Boolean CHR solver
as constraints, i.e. the disjunctions, are implicitly conjunctively
connected by the separating comma. For instance, the formula in
conjunctive normal form
\[
  (A \lor \lnot B \lor C) \land (\lnot A \lor B \lor D)
\]
is equivalent to 
\[
\mbox{\tt neg(A,F), neg(B,E), or(A,E,X), or(X,C,1), or(F,B,Y), or(Y,D,1)}
\]
if the semantics of the user-defined constraint {\tt neg(X,Y)} is
$\lnot X = Y$ and the semantics of the user-defined constraint {\tt
  or(X,Y,Z)} is $X \lor Y = Z$ for any arguments $X$, $Y$, and $Z$
that are either propositional variables, 0, or 1.  Thus, the important
class of SAT problems may be modelled as constraint problems and solved
by using the CHR handler with the following rules:
\\

\begin{minipage}[t]{0.38\textwidth}{\footnotesize
\begin{verbatim}
    or(0,X,Y) <=> Y=X.
    or(X,0,Y) <=> Y=X.
    or(X,Y,0) <=> X=0,Y=0.
    or(1,X,Y) <=> Y=1.
    or(X,1,Y) <=> Y=1.
    or(X,X,Z) <=> X=Z.
    neg(0,X) <=> X=1.
    neg(X,0) <=> X=1.
    neg(1,X) <=> X=0. 
    neg(X,1) <=> X=0.
    neg(X,X) <=> fail.
\end{verbatim}
}
\end{minipage}
\begin{minipage}[t]{0.58\textwidth}{\footnotesize
\begin{verbatim}
    or(X,Y,A) \ or(X,Y,B) <=> A=B.
    or(X,Y,A) \ or(Y,X,B) <=> A=B.
    neg(X,Y) \ neg(Y,Z) <=> X=Z.    
    neg(X,Y) \ neg(Z,Y) <=> X=Z.    
    neg(Y,X) \ neg(Y,Z) <=> X=Z.    
    neg(X,Y) \ or(X,Y,Z) <=> Z=1.
    neg(Y,X) \ or(X,Y,Z) <=> Z=1.
    neg(X,Z) , or(X,Y,Z) <=> X=0,Y=1,Z=1.
    neg(Z,X) , or(X,Y,Z) <=> X=0,Y=1,Z=1.
    neg(Y,Z) , or(X,Y,Z) <=> X=1,Y=0,Z=1.
    neg(Z,Y) , or(X,Y,Z) <=> X=1,Y=0,Z=1.
\end{verbatim}
}
\end{minipage}
\\
\\

The transformation of the CHR in the Java system~\cite{chrWolf:01a} is
straightforward:

\begin{example} 
  The coding of the first rule {\tt or(0,X,Y) <=> Y=X.} in a CHR
  handler is quite simple:
\begin{quote}{\footnotesize        
\noindent
\txt{class boolHandler} \BEGIN\\
(01)\tab \txt{public static void main(String[] args)} \BEGIN\\
(02)\tab\tab \txt{DJCHR djchr = new DJCHR("bool", new String[]\{"or/3","neg/2"\});}\\
(03)\tab\tab \txt{        Variable x = new Variable("X");}\\
(04)\tab\tab \txt{        Variable y = new Variable("Y");}\\
(05)\tab\tab \txt{        Term zero = new Term(0);}\\
\ldots\\ 
(06)\tab\tab \txt{        Term[] remove, keep, guard, body;}\\
\ldots\\
(07)\tab\tab \txt{        remove = new Term[]{new Term("or",new Term[]\{zero,x,y\})};}\\
(08)\tab\tab \txt{        body = new Term[]\{DJCHR.eq(y,x)\};}\\
(09)\tab\tab \txt{        djchr.addRule(remove,null,body,null);}\\
\ldots\\        
(10)\tab\tab \txt{        djchr.compileAll();}\\
(11)\tab \END\\
\END\\
}
\end{quote}
First of all, a new handler object {\tt djchr} is generated (line 2).
It is called {\tt bool} and supports the ternary user-defined
constraint {\tt or} and the binary user-defined constraint {\tt neg}.
Then, two variables {\tt X} and {\tt Y} as well as a constant 0 are
generated (lines 3--5). Every rule is split up into four arrays of
terms (line 6): the head constraints that are removed, the head
constraints that are kept, the guard constraints, and the body
constraints according to~\cite{chrHolzbaur:Fruehwirth:00a}. For the
considered rule to be transformed, the {\tt keep} and {\tt guard}
arrays must be empty. However, the {\tt remove} array contains the
constraint {\tt or(0,X,Y)}, which is generated accordingly (line 7).
Furthermore, the body constraint {\tt Y=X} is generated using of the
built-in method {\tt eq} (line 8).  Then the rule is composed and
added to the handler (line 9). Finally, all added rules are compiled
by calling {\tt compileAll()} (line 10). \qed
\end{example}
 
During the compilation process, a Boolean constraint solver class 
called {\tt bool} is generated.  The application interface generated for
this solver comprises the methods
\begin{itemize}
  \item {\tt void or\underline{~}3(Term[] args, IntegerSet set)}
  \item {\tt void neq\underline{~}2(Term[] args, IntegerSet set)}
\end{itemize}
to add the specified user-defined constraints {\tt or} and {\tt neg}.

Additionally, a class {\tt boolVariable} of \emph{attributed logical
  variables}, which is a subclass of the class {\tt Variable}, is also
generated. It has special attributes to store and access efficiently
the Boolean constraints on these variables, cf.~\cite{uHolzbaur:90,uWolf:01}.
Thus, the propositional formula in conjunctive normal form
\[
  (A \lor \lnot B \lor C) \land (\lnot A \lor B \lor D)
\]
is modelled as a Boolean constraint problem by the Java code fragment
\begin{quote}{\footnotesize
\noindent
\txt{bool cs = new bool();}\\
\txt{boolVariable a = new Variable("A");}\\
\txt{boolVariable b = new Variable("B");}\\
\txt{boolVariable c = new Variable("C");}\\
\txt{boolVariable d = new Variable("D");}\\
\txt{boolVariable e = new Variable("E");}\\
\txt{boolVariable f = new Variable("F");}\\
\txt{boolVariable x = new Variable("X");}\\
\txt{boolVariable y = new Variable("Y");}\\
\txt{Term one = new Term(1);}\\
\txt{cs.neg\underline{~}2(new Term[]\{a,f\}, $\emptyset$);}\\
\txt{cs.neg\underline{~}2(new Term[]\{b,e\}, $\emptyset$);}\\
\txt{cs.or\underline{~}3(new Term[]\{a,e,x\}, $\emptyset$);}\\ 
\txt{cs.or\underline{~}3(new Term[]\{x,c,one\}, $\emptyset$);}\\ 
\txt{cs.or\underline{~}3(new Term[]\{f,b,y\}, $\emptyset$);}\\ 
\txt{cs.or\underline{~}3(new Term[]\{y,d,one\}, $\emptyset$);}\\
}
\end{quote}
if it is assumed that the formula is always valid, which means that the
justifications are the empty sets. The calls of the methods {\tt
  cs.neg\underline{~}2} and {\tt cs.or\underline{~}3} add the
constraints to the constraint store of the constraint system {\tt cs} and
eventually trigger some of the compiled rules.

It should be noted that the presented Boolean CHR solver applies the
\emph{unit clause rule}~\cite{kDavisPutnam:60}. \emph{Unit clauses}
are disjunctions of literals, i.e. propositional variables or their
negations, where all literals except one are 0. Here, unit clauses are
represented by conjunctions of $k$~constraints 
\[
\mbox{{\tt or(X$_0$,X$_1$,R$_1$)}, {\tt or(R$_1$,X$_2$,R$_2$)}, \ldots, 
{\tt or(R$_k$,X$_k$,1)}}\enspace,
\]
where for a fixed index~$j \in \{1, \ldots, k\}$ it holds {\tt X}$_i = 0$ 
for all indices~$i \ne j$.

These constraints trigger the rule {\tt or(X,0,Y) <=> X=Y} several
times deriving in this order {\tt R$_k$ = \ldots\ = R$_j$ = 1}, and
further {\tt R$_1$ = \ldots\ = R$_{j-1}$} if $j>1$ holds. In any case,
either the rule {\tt or(X,0,Y) <=> X=Y} or {\tt or(0,X,Y) <=> X=Y} is
finally triggered, which results in {\tt X$_j$ = 1} in either case.

Other instances of propositional formulas in conjunctive
normal form that are processable using the introduced Boolean constraint
solver are the AIM instances presented in the next section.

\section{The AIM Instances}\label{aim}

The AIM instances are random 3-SAT problem instances in conjunctive
normal form, named after their originators Kazuo Iwama, Eiji Miyano
and Yuichi Asahiro. 3-SAT problems are conjunctions of disjunctions of
three literals, i.e. propositional variables or negations of them. The
AIM instances are all generated with a particular random 3-SAT
instance generator \cite{satAIM:96}. The particularity is that the
generator generates yes-instances and no-instances independently for
wide ranges. Thus its primary role is to provide the sort of instances
that conventional random generation has difficulty generating. The
generator runs in a randomised fashion, which means that the 3-SAT
instances essentially differ from those generated in a
deterministic fashion or from those translated from other problems.
As a result, the following set of considered AIM instances includes
\begin{itemize}
  \item no-instances with low clause/variable ratios that are inconsistent 
  \item yes-instances with low and high clause/variable ratios 
        that have exactly one solution
\end{itemize}
The instances are called {\tt aim-xxx-y\underline{~}y-zzzz-j} where
\begin{itemize}
  \item {\tt xxx} shows the number of variables, one of 50, 100 and 200
  \item {\tt y\underline{~}y} shows the clause/variable ratio {\tt y.y}, 
        including 1.6, 2.0 for no-instances and 1.6, 2.0, 3.4, and 6.0 
        for single-solution yes-instances 
  \item {\tt zzzz} is either ``no'' or ``yes1'', the former denoting a 
        no-instance and the latter a single-solution yes-instance
  \item the last {\tt j} means simply the {\tt j}-th instance at that 
        parameter
\end{itemize}
     
For each parameter, four instances are included in the benchmark
set. The whole benchmark set is available online at {\tt
http://www.satlib.org}.  For example, {\tt aim-50-1\underline{~}6-no-1}
through {\tt aim-50-1\underline{~}6-no-4} are four no-instances with
50~variables and a 1.6 clause/variable ratio. In all, there are 18
sets of instances with 50, 100 and 200 variables. For the 
yes-instances, clause /variable ratios are taken from 1.6, 2.0, 3.4, and
6.0; for the no-instances, they are taken from 1.6, and 2.0.

To find the unique solutions of the yes-instances or to prove the
inconsistency of the no-instances, there are several state-of-the-art
SAT solvers. A collection of SAT solvers is also available at {\tt
  http://www.satlib.org}. Most of these algorithms are (heuristic)
local-search algorithms or can be traced back to the Davis-Putnam
procedure~\cite{kDavisPutnam:60}. However, the presented Boolean
constraint solver in the previous section, complemented by a search
procedure that assigns the value 0 or 1 to the propositional variables,
can obviously be used to solve such SAT problems. Furthermore, SAT
problems are often used to compare ``intelligent'' search procedures
with chronological backtracking,
cf.~\cite{dcpBaker:94,dcpGinsberg:93,satLynce:MarquesSilva:02,dcpProsser:93}.
The next section therefore considers several search procedures and their
interaction with the generated Boolean constraint solver.

\section{The Search Procedures}\label{search}

Before describing the compared search procedures in detail, we look
at some of the assumptions made and programming conventions used.

\subsection{Programming Conventions}

It is assumed that there are Boolean Constraint Satisfaction Problems
(CSP), i.e. there are variables $V_1,\ldots,V_n$ with Boolean
domains $\{0,1\}$. Additionally, there are two types of Boolean
constraints over these variables, either negations $\lnot X = Y$ or
disjunctions $X \lor Y = Z$, where $X$, $Y$ or $Z$ are either
variables, 0 or 1. The problem is either to detect that there is no
assignment of values to the variables such that the constraints are
satisfied, i.e. the problem is inconsistent, or to find such an
assignment, i.e. a solution. The Boolean constraints are realized by
user-defined constraints handled by the Boolean solver presented in
Section~\ref{boolSolver}.

The different search procedures to solve Boolean CSP are presented in
pseudo-code strongly related to Java. The main difference compared to
Java is that mathematical set notation is used instead of some methods
of the ``abstract'' class {\tt IntegerSet}. -- Actually, we used
our implementation of sparse integer sets, which is described
in~\cite{chrWolf:99d}.  However, this might be replaced by any other,
even more efficient implementation.

It is assumed that there is a globally declared array of
variables\footnote{Variables in the sense of constraint processing.}
{\tt var}, such that {\tt var[i]} represents the variable $V_i$ for
$i=1,\ldots,n$ where $n$ is the actual number of variables in
the considered problem.  Variables (of the class {\tt Variable})
implement attributed logical variables: they may be bound to
terms, e.g. integers, or unbound, i.e. free. Thus, there is the
method
\begin{itemize}
  \item {\tt boolean isBound()} which returns true if and only if 
        the variable is bound.
\end{itemize}
If a variable is bound, the method {\tt Term value()} is defined, which
returns the term the variable is bound to. Furthermore, there is an
integer field {\tt num} holding either the next value to be assigned
to this variable (see Section~\ref{secCBT}) or an identifier justifying 
the current assignment (see Section~\ref{secDBT}).

A variable also contains an array of integer sets with indices ranging
over the Boolean domain from 0 to 1.  If defined, i.e. if different
from {\tt null}, this array contains for each value unique identifiers
of the variables, i.e. their indices, bound to values that result in
an inconsistency, which was detected with respect to the considered
Boolean constraint problem.

\begin{example}
Let this set for the value 1 of the variable $V_{17}$ be $\{3,7,8,10\}$
where 3,7,8, and 10 are the indices of other labelled variables. Then the
assignment $V_{17} = 1$ is inconsistent with the current assignments to
the variables $V_3$, $V_7$, $V_8$ and $V_{10}$ with respect to the
considered Boolean CSP. \qed
\end{example}

In conflict-directed backjumping, these sets are called \emph{conflict
  sets}, and in dynamic backtracking they are called \emph{elimination
  explanations}.  Thus, in these search procedures the array is
declared as either
\begin{itemize}
  \item {\tt IntegerSet[] conflictSet} or
  \item {\tt IntegerSet[] elimExpl},
\end{itemize}
accordingly. Furthermore, it is assumed that the language
supports variable lists, e.g.  a ``wrapper'' {\tt VariableList} of the
Java class {\tt ArrayList} that supports
\begin{itemize} 
  \item access to the size of a list: {\tt int size()}
  \item addition of a variable at the end of a list: 
        {\tt void add(Variable var)}
  \item access to a variable at a specific position in a list: 
        {\tt Variable get(int i)}, \\
        where the index of the first variable in a list is zero
  \item access to the last variable in a list: 
        {\tt Variable getLast(int i)} 
  \item removal of a variable at a specific position: 
        {\tt Variable remove(int i)}\\ 
        such that the indices of the variables that come after the 
        removed variable are decremented by one
\end{itemize}
The following data structures are also assumed to be globally declared
and thus accessible to all methods:
\begin{itemize}
  \item A unique Boolean constraint system {\tt cs} of the class {\tt bool},
        where the constraints are stored and processed by use of the 
        Boolean CHR solver (see Section~\ref{boolSolver}).
  \item Two variable lists {\tt unlabelledVars} and {\tt labelledVars} 
        of the class {\tt VariableList}, where the unlabelled and labelled 
        variables are stored during dynamic backtracking 
        (cf. Figures~\ref{dbtLabel} and~\ref{dbtUnlabel}).
  \item A unique integer {\tt cntr}, which is initially 0 and incremented 
        by one after an assignment in dynamic backtracking 
        (cf. Figure~\ref{dbtLabel}, line 10) serving as its unique 
        justification. 
\end{itemize}
In the sequel, the calls to the constraint system using the interface
to the adaptive CHR system are underlined.  This shows the simple and
powerful use of our adaptive CHR system in sophisticated search
procedures.

\subsection{The Constraint Satisfaction Search Process}

According to the style presented in~\cite{dcpProsser:93}, the
\emph{constraint satisfaction search problem} ({\tt cssp}) method in
Figure~\ref{cssp} establishes the environment in which the
different search methods are called. The  {\tt cssp} method takes the 
total number of variables to be labelled with values and returns
true if a solution is found and false if the given Boolean CSP is
inconsistent.
\begin{figure}
\begin{quote}
\txt{static boolean cssp(int n)} \BEGIN \\
(01)\tab \txt{int i = 1};\\
(02)\tab \WHILE \txt{(1 <= i \&\& i <= n)} \BEGIN\\
(03)\tab\tab \txt{int j = xxxLabel(i)};\\
(04)\tab\tab \IF \txt{(i == j)}\\
(05)\tab\tab\tab \txt{i = xxxUnlabel(i)};\\
(06)\tab\tab \ELSE \txt{i = j};\\
(07)\tab \END\\
(08)\tab \IF \txt{(i = 0)}\\
(09)\tab\tab \RETURN \txt{false};\\
(10)\tab \IF \txt{(i > n)}\\
(11)\tab\tab \RETURN \txt{true};\\
\END
\end{quote}
\caption{The {\tt cssp} function for solving constraint satisfaction 
         search problems}\label{cssp}
\end{figure}
The ``generic'' methods {\tt xxxLabel} and {\tt xxxUnlabel} are
replaced in the sequel resulting in chronological backtracking
({\tt cbtLabel/cbtUnlabel}), conflict-directed backjumping ({\tt
  cbjLabel/cbjUnlabel}) and two variants of dynamic backtracking ({\tt
  dbtLabel/dbtUnlabel} and {\tt fbtLabel/fbtUnlabel}). In all these
instances, the method {\tt xxxLabel} attempts to find a consistent
assignment to the $i$-th variable.\footnote{The $i$-th variable
  coincides with~$V_i$ in chronological backtracking and
  conflict-directed backjumping but not necessarily in dynamic
  backtracking, which may dynamically change the variable ordering.}
For this, the method takes $i$ as its argument. It returns this given
integer if no such assignment is found. However, if a consistent
assignment to the $i$-th variable is found, it returns $i+1$ after
binding this variable to a value that is consistent with the other
$i-1$ previously bound variables and with respect to the given Boolean
constraint problem. If {\tt xxxLabel} returns $i$, the
method {\tt xxxUnlabel} is called. When $i+1$ is returned with $1 <
i+1\le n$, {\tt xxxLabel} is called again, looking for an assignment to
the $(i+1)$-th variable. Returning $n+1$ causes {\tt cssp} to return
{\tt true} because a consistent assignment for all variables is found.

The corresponding instance of {\tt xxxUnlabel} is called when no
consistent assignment to the $i$-th variable is found (cf. lines 4--5
in Figure~\ref{cssp}). It performs backtracking from the $i$-th
variable to an $h$-th variable ($h<i$) if another value for the $h$-th
variable might resolve the inconsistency detected at the $i$-th variable. 
It takes $i$ as its argument. It either returns 0 or the index of the next
variable to be labelled. Zero is returned if the detected inconsistency
is not resolvable, i.e. the given Boolean CSP is inconsistent causing
{\tt cssp} to return {\tt false} ( Figure~\ref{cssp}, lines 8--9).

\subsection{Chronological Backtracking}\label{secCBT}

\begin{floatingfigure}{0.35\textwidth}
\hspace*{-0,05\textwidth}
\epsfig{file=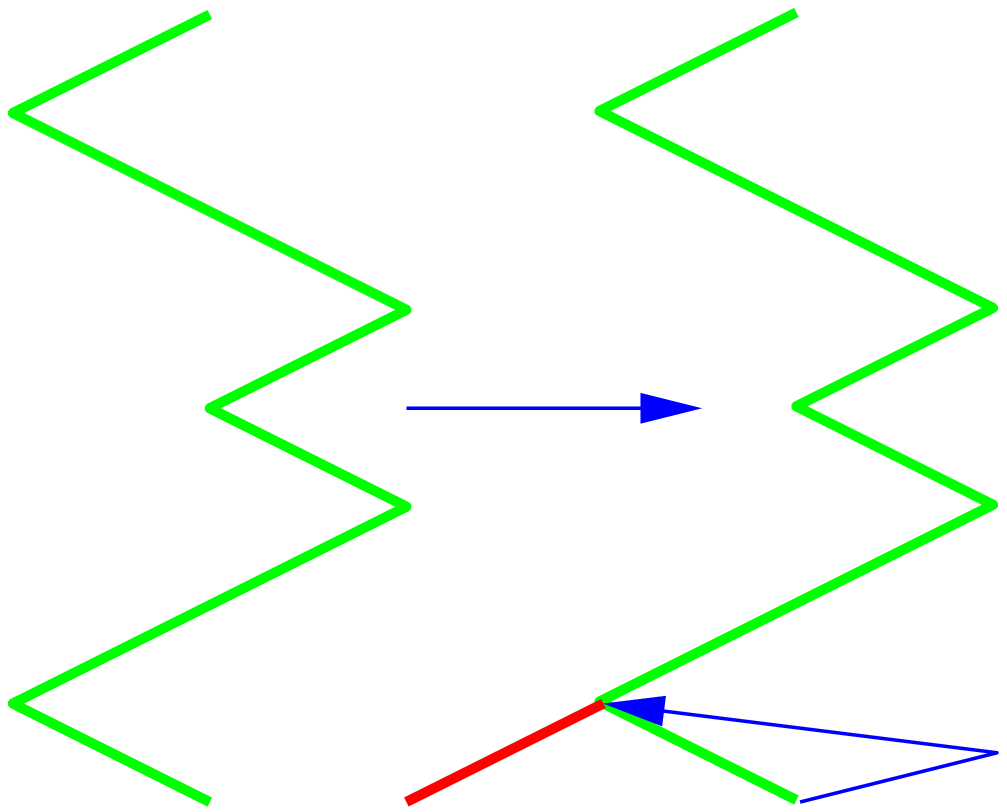, width=0.35\textwidth}
\caption{The principle of chronological backtracking}\label{cbt}
\end{floatingfigure}

\noindent
Chronological backtracking (CBT) is a simple depth-first search (cf.
Figure~\ref{cbt}) with a fixed tree structure, i.e. variable ordering.
If the variables are not already bound by constraint processing
(Figure~\ref{cbtLabel}, lines 1--2), they are incrementally bound
to the values 0 or 1.  First, the current variable $V_i$ is labelled
with the value 0. Search continues with $V_{i+1}$ if no inconsistency is
detected (Figure~\ref{cbtLabel}, lines 5--6). Otherwise, the value 1 is
assigned to the variable $V_i$. Again, search continues with $V_{i+1}$ if no
inconsistency is detected. Otherwise, a dead end is reached and the
search backtracks to the variable $V_{i-1}$ (cf.  Figures~\ref{cbt}
and~\ref{cbtUnlabel}).

\begin{figure}[hb]
\begin{quote}
\txt{int cbtLabel(int i)} \BEGIN \\
(01)\tab \IF \txt{(var[i].isBound())} \\
(02)\tab\tab  \RETURN \txt{i+1;}\\
(03)\tab \WHILE \txt{(var[i].num <= 1)} \BEGIN \\
(04)\tab\tab \txt{\underline{cs.equal(var[i], var[i].num++, $\{i\}$)};}\\
(05)\tab\tab \IF \txt{(\underline{cs.isConsistent()})} \\
(06)\tab\tab\tab \RETURN \txt{i+1;} \\
(07)\tab\tab \ELSE \\
(08)\tab\tab\tab \txt{\underline{cs.delete($\{i\}$)};}\\
(09)\tab \END\\
(10)\tab \RETURN \txt{i;}\\
\END
\end{quote}
\caption{The labelling method of chronological backtracking}\label{cbtLabel}
\end{figure}

A generalisation of this search process for arbitrary finite domains
is quite simple: the field {\tt num} in the variable must be replaced
by the domain.  During labelling, it must be iterated over the values
in the current domain (Figure~\ref{cbtLabel}, lines 3--9). The iterator for
this loop must be reset during unlabelling (Figure~\ref{cbtUnlabel},
line 2).\\
 
\begin{figure}
\begin{quote}
\txt{int cbtUnlabel(int i)} \BEGIN \\
(01)\tab \txt{\underline{cs.delete($\{i\}$)};}\\
(02)\tab \txt{var[i].num=0;}\\
(03)\tab \RETURN \txt{i-1;}\\
\END
\end{quote}
\caption{The unlabelling method of chronological backtracking}\label{cbtUnlabel}
\end{figure}

\subsection{Conflict-directed Backjumping}\label{secCBJ}

Conflict-directed backjumping (CBJ)~\cite{dcpProsser:93} is a guided
depth-first search with a fixed tree structure, i.e. variable ordering,
that ``jumps back'' to the most recent variable assignment that is in
conflict with the current variable (cf. Figure~\ref{cbj}). Originally, 
CBJ maintains a conflict set per variable. However, in our refinement
it maintains a conflict set for each value of every variable. Initially,
these conflict sets are not defined, i.e. {\tt null}.

\begin{floatingfigure}{0.35\textwidth}
\hspace*{-0,025\textwidth}
\epsfig{file=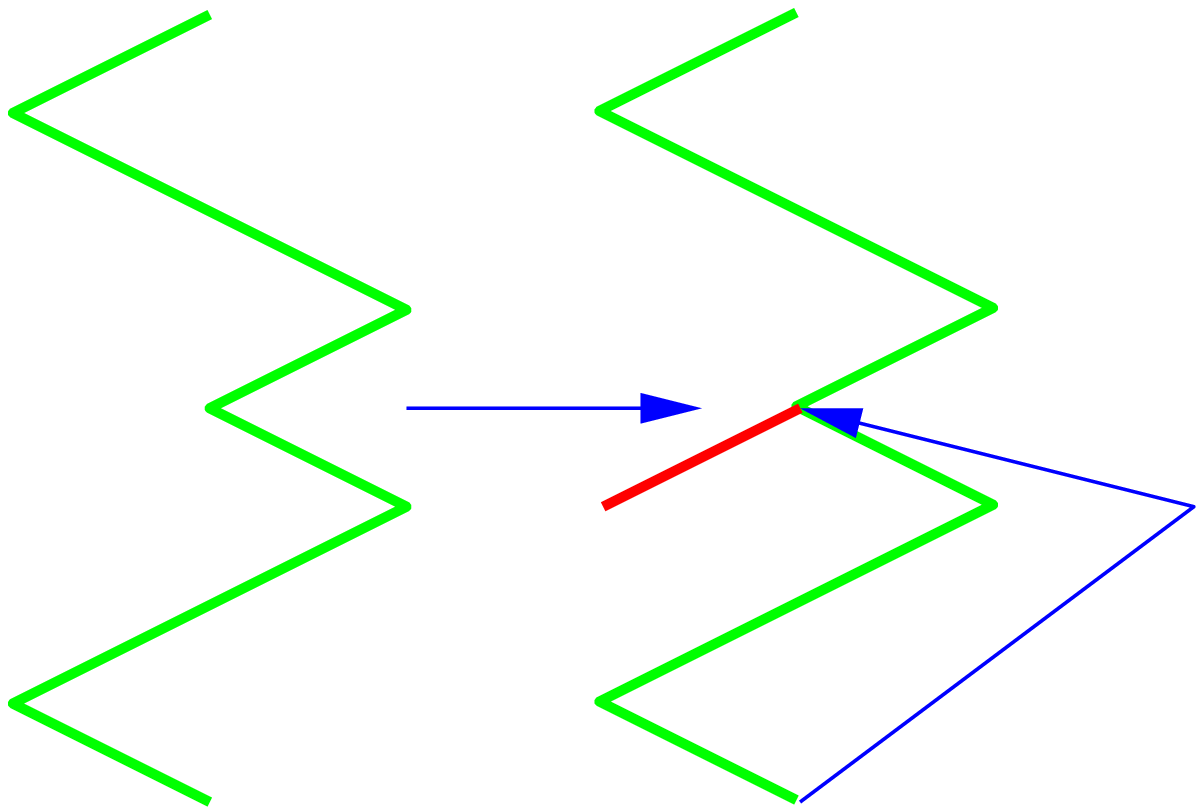, width=0.35\textwidth}
\caption{The principle of conflict-directed backjumping}\label{cbj}
\end{floatingfigure}

If the unlabelled variable $V_i$ is not already bound by constraint
processing (Figure~\ref{cbjLabel}, lines 1--2) the attempt is made to
bind it either to the value 0 or 1.  The current variable $V_i$ is
labelled with the first value that is possibly not in conflict with
other already labelled variables (Figure~\ref{cbjLabel}, line 4--5).
Thus, the index of the variable is chosen as the justification of this
assignment because it simply allows any subsequent deletion of it and
all its consequences computed by the underlying Boolean constraint
solver (cf.  Figure~\ref{cbjLabel}, line 10 and
Figure~\ref{cbjUnlabel}, line 6).

The search continues with $V_{i+1}$ if no inconsistency is detected
(Figure~\ref{cbjLabel}, lines 6--7).  Otherwise, the indices of the
already labelled variables that are responsible for the detected
inconsistency form the conflict set of the attempted value
(Figure~\ref{cbjLabel}, lines 8--11), the assignment is deleted
(Figure~\ref{cbjLabel}, line 10) and the next value for $V_i$ is
attempted (Figure~\ref{cbjLabel}, lines 3--13). If all assignments
lead to an inconsistency, a dead end is reached, i.e. $i$ is returned
(Figure~\ref{cbjLabel}, line 14), which triggers unlabelling.  If the
conflict sets of all values of the considered variable $V_i$ are
empty, the deletion of all variable assignments will not resolve the
detected inconsistency. Thus, 0 is returned, indicating the
inconsistency of the given Boolean CSP (Figure~\ref{cbjUnlabel}, lines
1--2).  If the union of all conflict sets is not empty, the search
``jumps back'' to the most recent assignment that is involved in the
detected dead end. This means that the assignment to the variable
$V_{h}$ is involved in the reached dead end, where $h$ is the largest
index in this union (cf. Figure~\ref{cbjUnlabel}, line 3).  Before
jumping back, the not yet defined conflict set of the value assigned
to the variable~$V_h$ (cf.  Figure~\ref{cbjLabel}, line 4) becomes the
union of the conflict sets of the values attempted for the
variable~$V_i$ without the index $h$ (Figure~\ref{cbjUnlabel}, lines
4--5). This is crucial because without any change in the assignments
to the variables indicated in this conflict set, the variables
from~$V_h$ to $V_i$ will be bound to the same values leading to the
same dead end, and thus into a loop.  Then the assignments to the
variables from~$V_h$ to $V_{i-1}$ are deleted
(Figure~\ref{cbjUnlabel}, line 6) and the conflict sets of all
previously labelled variables ($V_h$ to $V_n$) are updated, i.e.  all
defined conflict sets indicating variables that are not deleted are
kept because they are still valid (Figure~\ref{cbjUnlabel}, lines
7--15). Finally, $h$, the index of the next variable to be labelled,
is returned (Figure~\ref{cbjUnlabel}, line 16).

The method proposed here is in several respects more general than the
original CBJ or its extensions with forward checking (FC-CBJ) also
presented in~\cite{dcpProsser:93}, or with maintaining arc consistency
(MAC-CBJ) presented in~\cite{dcpProsser:95}: 

Firstly, our algorithm is not restricted to binary constraints;
it processes constraints of arbitrary arities.  Secondly, instead of
checking each assignment of the current variable against the
assignments to the already bound variables to determine the conflict
sets as in the original CBJ, constraint propagation is used in our
approach to detect inconsistencies and their explanations.  This is
similar to MAC-CBJ~\cite{dcpProsser:95}, where constraint propagation
performs arc consistency. However, the underlying CHR solver is able
to perform stronger, more ``global'' propagation because multi-headed
rules allow reasoning over combinations of several constraints:

\begin{example}\label{mac-expl}
  The single-headed rules of the Boolean CHR solver introduced in
  Section~\ref{boolSolver} perform local propagation maintaining local
  consistency (cf.~\cite{cMarriott:Stuckey:98}), which is the canonical
  extension of arc consistency to non-binary constraint problems.
  Furthermore, the two-headed rules perform additional propagation:
  
  Given: the Boolean variables {\tt U}, {\tt V}, {\tt X}, and {\tt Y}
  with domains $\{0,1\}$ as well as the constraints {\tt or(X,U,V)},
  {\tt neg(Y,U)}, and {\tt or(X,Y,V)}. In the first search step, we
  label {\tt X = 0}.  The original CBJ is unable to perform at all
  because all constraints have unbound variables. Neither forward
  checking in FC-CBJ nor MAC-CBJ will restrict any domains of the
  not-yet-labelled variables. However, in our approach this labelling
  triggers the rule {\tt or(0,U,V) <=> U=V}. This simplifies the
  constraints to {\tt U = V}, {\tt neg(Y,V)} and {\tt or(X,Y,U)}. The
  equation {\tt U=V} further triggers the rule {\tt neg(Y,V),
    or(X,Y,V) <=> X=1, Y=0, V=1} resulting in an inconsistency.
\end{example}

Thus, in our approach the assignment to the current variable is not
only checked against past variable assignments but also against the
constraints with future variables, maintaining some kind of
consistency that is in general stronger than local consistency.

As aforementioned, the conflict sets in our approach are not only
stored for each variable, they are stored for all possible values of
each variable also proposed by~\cite{lpBruynooghe:04}. This allows us
to avoid already detected conflicts after any back-jumps or
re-assignments to variables:

\begin{example}
  Let us assume that the value~0 of the variable $V_7$ is in conflict
  with the assignments to the variables $V_1$ and $V_3$ and the value
  1 of the variable $V_7$ is in conflict with the assignments to the
  variables $V_2$ and $V_4$.  Then, CBJ jumps back to the variable
  $V_4$, undoing the assignments to the variables $V_7$, $V_6$, $V_5$
  and $V_4$. The value recently assigned to the variable $V_4$ is thus
  known to be in conflict with the variables $V_1$, $V_2$ and $V_3$.
  If there is any non-conflicting assignment to the variable $V_4$, we
  know for future labelling that the value~0 of the variable $V_7$ is
  still in conflict with the assignments to the variables $V_1$ and
  $V_3$. \qed
\end{example} 

A generalisation of this search process for arbitrary finite domains
is quite simple: It must be iterated over the values in the domain
(Figure~\ref{cbjLabel}, lines 3--13 and Figure~\ref{cbjUnlabel}, lines
9--15), and the tests and calculations must be done for all
conflict sets of the domain values ( Figure~\ref{cbjUnlabel}, lines 1
and 3--5).
 
\begin{figure}
\begin{quote}
\txt{int cbjLabel(int i)} \BEGIN \\
(01)\tab \IF \txt{(var[i].isBound())} \\
(02)\tab\tab  \RETURN \txt{i+1;}\\
(03)\tab \FOR \txt{(int k=0; k <= 1; k++)} \BEGIN \\
(04)\tab\tab \IF \txt{(var[i].conflictSet[k] == null)} \BEGIN \\
(05)\tab\tab\tab \txt{\underline{cs.equal(var[i], k, \{i\})};}\\
(06)\tab\tab\tab \IF \txt{(\underline{cs.isConsistent()})} \\
(07)\tab\tab\tab\tab \RETURN \txt{i+1;} \\
(08)\tab\tab\tab \ELSE \BEGIN \\
(09)\tab\tab\tab\tab \txt{var[i].conflictSet[k] = \underline{cs.getExplanation()}$\setminus$\{i\};} \\
(10)\tab\tab\tab\tab \txt{\underline{cs.delete(\{i\})};}\\
(11)\tab\tab\tab \END \\
(12)\tab\tab \END\\
(13)\tab \END\\
(14)\tab \RETURN \txt{i;}\\
\END
\end{quote}
\caption{The labelling method of conflict-directed backjumping}\label{cbjLabel}
\end{figure}

\begin{figure}
\begin{quote}
\txt{int cbjUnlabel(int i)} \BEGIN \\
(01)\tab \IF \txt{(var[i].conflictSet[0] == $\emptyset$ \&\& var[i].conflictSet[1] == $\emptyset$)}\\
(02)\tab\tab \RETURN \txt{0;}\\
(03)\tab \txt{int h = $\max(\txt{var[i].conflictSet[0]} \cup \txt{var[i].conflictSet[1]})$}\\
(04)\tab \txt{var[h].conflictSet[var[h].value()]}\\
(05)\tab\tab \txt{= (var[i].conflictSet[0] $\cup$ var[i].conflictSet[1])$\setminus\{h\}$;}\\
(06)\tab \txt{\underline{cs.delete($\{h,\ldots,i-1\}$)};}\\
(07)\tab \FOR \txt{(int j=h; j <= n; j++)} \BEGIN\\
(08)\tab\tab \FOR \txt{(int k=0; k <= 1, k++)} \BEGIN \\
(09)\tab\tab\tab \IF \txt{(var[j].conflictSet[k] != null}\\
(10)\tab\tab\tab\tab\tab \txt{\&\& var[j].conflictSet[k] != $\emptyset$}\\ 
(11)\tab\tab\tab\tab\tab \txt{\&\& $\max(\txt{var[j].conflictSet[k]})$ >= h)} \BEGIN\\
(12)\tab\tab\tab\tab \txt{ var[j].conflictSet[k] = null;}\\ 
(13)\tab\tab\tab \END\\
(14)\tab\tab \END\\
(15)\tab \END\\
(16)\tab \RETURN \txt{h;}\\
\END\\
\end{quote}
\caption{The unlabelling method of conflict-directed backjumping keeping formerly detected and still valid conflict sets}\label{cbjUnlabel}
\end{figure}

\subsection{Dynamic Backtracking}\label{secDBT}

\begin{floatingfigure}{0.35\textwidth}
\hspace*{-0,025\textwidth}
\epsfig{file=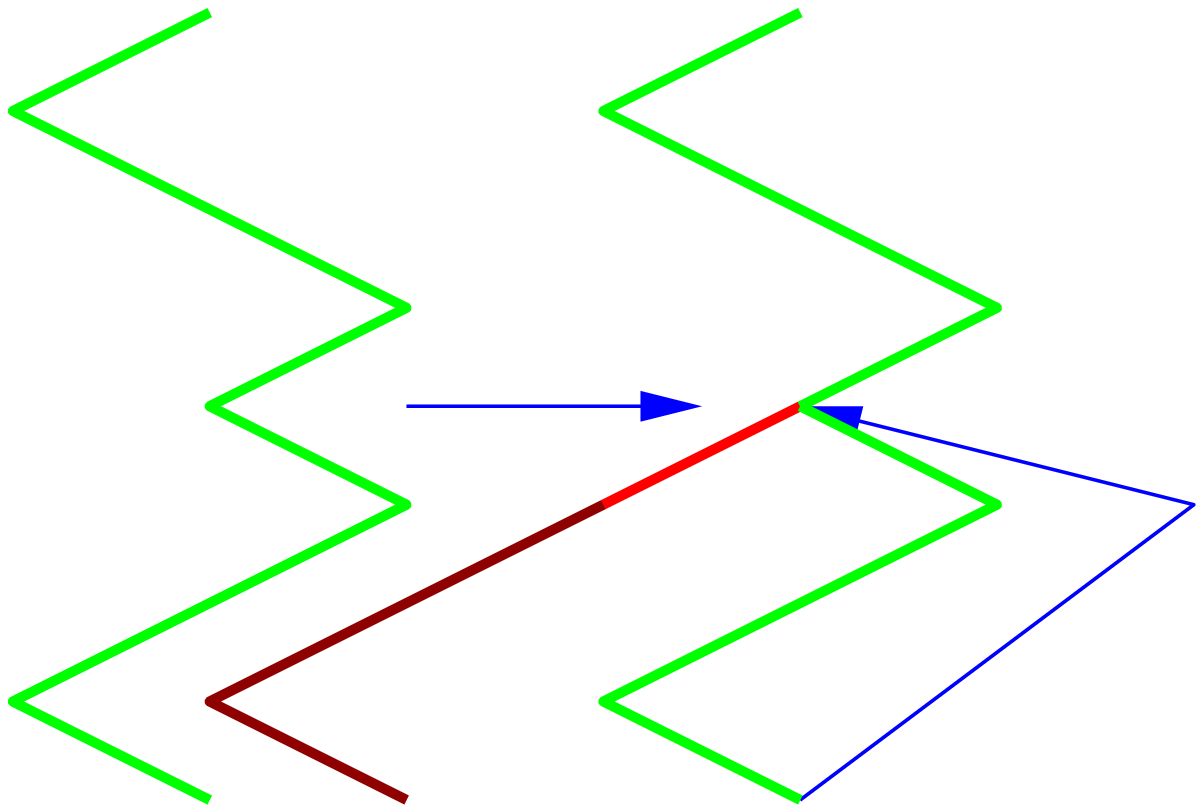, width=0.35\textwidth}
\caption{The principle of dynamic backtracking}\label{dbt}
\end{floatingfigure}

Dynamic backtracking (DBT)~\cite{dcpGinsberg:93} is a guided
depth-first search dynamically changing the tree structure, i.e. the
variable ordering, which goes back to the most recent variable
assignment that is in conflict with the current variable retaining the
intermediate assignments (cf.  Figure~\ref{dbt}). DBT maintains an
elimination explanation for each value of every variable. Initially,
these sets are not defined, i.e. {\tt null}. In DBT, two global
variable lists are maintained to manage the dynamic changes of the
value ordering.  The list {\tt unlabelledVars} contains the
not-yet-labelled variables, while the list {\tt labelledVars} keeps
the already labelled variables.

If the (last-entered) unlabelled variable is not already bound by
constraint processing (Figure~\ref{dbtLabel}, lines 1--5) the attempt
is made to bind it either to the value 0 or 1.  This variable is 
labelled with the first value that is possibly not in conflict with
other already labelled variables (Figure~\ref{dbtLabel}, line 7--8).
Thus, the value of the global counter is chosen as its unique
justification and later stored at the variable if no inconsistency
arises (Figure~\ref{dbtLabel}, lines 8 and 10). This facilitates any
subsequent deletion of the assignment and all its consequences computed by
the underlying Boolean constraint solver (cf.  Figure~\ref{dbtLabel},
line 16 and Figure~\ref{dbtUnlabel}, line 15).

The search continues with the next unlabelled variable if no
inconsistency is detected (Figure~\ref{dbtLabel}, lines 10--12).
Otherwise, the justifications of the already labelled variables that are
responsible for the detected inconsistency form the elimination
explanation of the attempted value (Figure~\ref{dbtLabel}, line 15),
the assignment is deleted (Figure~\ref{dbtLabel}, line 16), and the
next value for this variable is attempted (Figure~\ref{dbtLabel},
lines 6--19). If each assignment leads to an inconsistency, a dead end
is reached, i.e. the variable is added to the list of unlabelled
variables and $i$ is returned (Figure~\ref{dbtLabel}, lines 20--21),
which triggers unlabelling.  If the elimination explanation of all
values of the considered variable are empty, the deletion of all
variable assignments will not resolve the detected inconsistency.
Thus, 0 is returned, indicating the inconsistency of the given Boolean
CSP (Figure~\ref{dbtUnlabel}, lines 1--2).  If the union of all 
elimination explanations is not empty, the search ``goes back'' to the
most recent assignment that is involved in the detected dead end.
This means that the assignment to the variable {\tt bt} justified by
the maximum~{\tt h} in this union (cf. Figure~\ref{dbtUnlabel}, lines
5--12) is involved in the reached dead end.  Before going back, the
elimination explanation of the value assigned to the variable~{\tt bt}
becomes the union of the elimination explanations of the values
attempted for the most recently tried variable in {\tt dbtLabel}.
This causes the justification $h$ (Figure~\ref{dbtUnlabel}, lines
13--14) to be removed. This is crucial as in CBJ (see
Section~\ref{secCBJ}) because otherwise {\tt bt} will be labelled again
with the same value leading to the same dead end, and thus into a
loop.  The assignment to the variable {\tt bt} is then deleted
(Figure~\ref{dbtUnlabel}, line 15) and added to the unlabelled
variables. Additionally, the elimination explanations of all variables
are updated, i.e. all defined elimination explanations not containing
the justification of the deleted assignment are kept because they are
still valid (Figure~\ref{dbtUnlabel}, lines 17--33).

\begin{example}
  Let us assume that the elimination explanation of the value~0 of the
  variable $V_7$ consists of the justifications of the assignments to
  the variables $V_2$ and $V_9$ and that the elimination explanation of the
  value 1 of the variable $V_7$ consists of the justifications of the
  assignments to the variables $V_3$ and $V_{11}$. The most recent
  assignment (with the largest justification) is assumed to be to the
  variable~$V_3$. Thus, DBT goes back to the variable $V_3$, undoing
  its assignment. The value recently assigned to the variable $V_3$ is
  thus known to be in conflict with the assignments to the variables
  $V_2$, $V_9$ and $V_{11}$. Thus, the elimination explanation of
  this value is the union of the justifications of these variables. If
  there is any non-conflicting assignment to the variable $V_3$, we
  know for future labelling that the value~0 of the variable $V_7$ is
  still in conflict with the assignments to the variables $V_2$ and
  $V_9$. Furthermore, any other elimination explanation not containing
  the justification of the removed assignment is still valid. \qed
\end{example} 

As a result of the non-chronological constraint deletion, a previously
bound variable that is stored in the list of already labelled
variables (cf. Figure~\ref{dbtLabel}, lines 2--3) may happen to be
unbound. Such free variables are filtered out and moved to the
variables that still have to be labelled (cf. Figure~\ref{dbtUnlabel},
lines 34--37).

Finally, the number of the variable that has to be labelled next is
returned (Figure~\ref{dbtUnlabel}, line 39).\footnote{Note that the
  number is not necessarily the index in the array {\tt var} because
  the value ordering may change dynamically.}

The method proposed here is in several respects more general than the
original DBT~\cite{dcpGinsberg:93} or its extensions with forward 
checking (FC-DBT) or even with maintaining arc consistency
(MAC-DBT) presented in~\cite{dcpJussien:etal:00}: 

Firstly, our algorithm is not restricted to binary constraints;
it processes constraints of arbitrary arities.  Secondly, instead of
checking each assignment of the current variable against the
assignments to the already bound variables to determine the elimination
explanation as in the original DBT, constraint propagation is used in our
approach to detect inconsistencies and their explanations.  This is
similar to MAC-DBT~\cite{dcpJussien:etal:00}, where constraint propagation
performs arc consistency. However, the underlying CHR solver is able
to perform stronger, more ``global'' propagation because multi-headed
rules allow reasoning over combinations of several constraints 
(cf.~Example~\ref{mac-expl}).

Unlike other solvers for dynamic CSP~\cite{dcpJussien:etal:00} our
underlying adaptive CHR constraint solver which was primarily
constructed to solve dynamic CSP, is adequate to support DBT for
dynamic CSP~\cite{dcpVerfaillie:Schiex:94a}: justifications are not
restricted to variable assignments; any other constraint may be
justified, too. Thus, the crucial calculation of elimination
explanations performed by the method {\tt getExplanation()} returns
the identifiers of the \emph{constraints} involved in the detected 
inconsistency.

A generalisation of this search process for arbitrary finite domains
is quite simple: It must be iterated over the values in the domain
(Figure~\ref{dbtLabel}, lines 6--19 and Figure~\ref{dbtUnlabel}, lines
19--24 and 28--33), and the tests and calculations must be done
for all eliminations explanation of the domain values
(Figure~\ref{cbjUnlabel}, lines 2 and 6).

\begin{figure}
\begin{quote}
\txt{int dbtLabel(int i)} \BEGIN \\
(01)\tab \txt{Variable var = unlabelledVars.removeLast();}\\
(02)\tab \IF \txt{(var.isBound())} \BEGIN\\
(03)\tab\tab \txt{labelledVars.add(var);}\\
(04)\tab\tab \RETURN \txt{i+1;}\\
(05)\tab \END\\
(06)\tab \FOR \txt{(int k=0; k <= 1; k++)} \BEGIN \\
(07)\tab\tab \IF \txt{(var.elimExpl[k] == null)} \BEGIN \\
(08)\tab\tab\tab \txt{\underline{cs.equal(var, k, \{cntr\})};}\\
(09)\tab\tab\tab \IF \txt{(\underline{cs.isConsistent()})} \BEGIN \\
(10)\tab\tab\tab\tab \txt{var.num = cntr++;}\\
(11)\tab\tab\tab\tab \txt{labelledVars.add(var);}\\
(12)\tab\tab\tab\tab \RETURN \txt{i+1;} \\
(13)\tab\tab\tab \END \\
(14)\tab\tab\tab \ELSE \BEGIN\\
(15)\tab\tab\tab\tab \txt{var.elimExpl[k] = \underline{cs.getExplanation()}$\setminus$\{cntr\};} \\
(16)\tab\tab\tab\tab \txt{\underline{cs.delete(\{cntr\})};}\\
(17)\tab\tab\tab \END\\
(18)\tab\tab \END\\
(19)\tab \END\\
(20)\tab \txt{unlabelledVars.add(var);}\\
(21)\tab \RETURN \txt{i;}\\
\END
\end{quote}
\caption{The labelling method of dynamic backtracking}\label{dbtLabel}
\end{figure}

\begin{figure}
\begin{quote}
\txt{int dbtUnlabel(int i)} \BEGIN \\
(01)\tab \txt{Variable var = unlabelledVars.removeLast();}\\
(02)\tab \IF \txt{(var.elimExpl[0] == $\emptyset$ \&\& var.elimExpl[1] == $\emptyset$)} \BEGIN\\
(03)\tab\tab \RETURN \txt{0;}\\
(04)\tab \END\\
(05)\tab \txt{Variable bt;}\\
(06)\tab \txt{int h = $\max(\txt{var.elimExpl[0]} \cup \txt{var.elimExpl[1]})$;}\\
(07)\tab \FOR \txt{(int j=labelledVars.size()-1; j >= 0; j--)} \BEGIN\\
(08)\tab\tab \txt{bt = labelledVars.get(j);}\\
(09)\tab\tab \IF \txt{(bt.num == h)} \BEGIN\\
(10)\tab\tab\tab \txt{labelledVars.remove(j);}\\
(11)\tab\tab\tab \txt{break;}\\
(12)\tab\tab \END\\
(13)\tab \txt{bt.elimExpl[bt.value()]}\\
(14)\tab\tab \txt{= (var.elimExpl[0] $\cup$ var.elimExpl[1])$\setminus\{h\}$;}\\
(15)\tab \txt{\underline{cs.delete($\{h\}$)};}\\
(16)\tab \txt{unlabelledVars.add(bt);}\\
(17)\tab \FOR \txt{(j=0; j < unlabelledVars.size(); j++)} \BEGIN\\
(18)\tab\tab \txt{var = unlabelledVars.get(j);}\\
(19)\tab\tab \FOR \txt{(int k=0; k <= 1, k++)} \BEGIN \\
(20)\tab\tab\tab \IF \txt{(var.elimExpl[k] != null}\\
(21)\tab\tab\tab\tab\tab \txt{\&\& h $\in$ var.elimExpl[k])} \BEGIN\\ 
(22)\tab\tab\tab\tab \txt{ var.elimExpl[k] = null;}\\ 
(23)\tab\tab\tab \END\\
(24)\tab\tab \END\\
(25)\tab \END\\
(26)\tab \FOR \txt{(j=labelledVars.size()-1; j >=0; j--)} \BEGIN\\
(27)\tab\tab \txt{var = labelledVars.get(j);}\\
(28)\tab\tab \FOR \txt{(int k=0; k <= 1, k++)} \BEGIN \\
(29)\tab\tab\tab \IF \txt{(var.elimExpl[k] != null}\\
(30)\tab\tab\tab\tab\tab \txt{\&\& h $\in$ var.elimExpl[k])} \BEGIN\\ 
(31)\tab\tab\tab\tab \txt{ var.elimExpl[k] = null;}\\ 
(32)\tab\tab\tab \END\\
(33)\tab\tab \END\\
(34)\tab\tab \IF \txt{(!var.isBound())} \BEGIN\\
(35)\tab\tab\tab \txt{labelledVars.remove(j);}\\ 
(36)\tab\tab\tab \txt{unlabelledVars.add(var);}\\ 
(37)\tab\tab \END\\
(38)\tab \END\\
(39)\tab \RETURN \txt{labelledVars.size()+1;}\\
\END\\
\end{quote}
\caption{The unlabelling method of dynamic backtracking keeping formerly 
  detected and still valid elimination explanations}\label{dbtUnlabel}
\end{figure}

\subsection{``Fancy'' Backtracking}

The variant of dynamic backtracking presented in~\cite{dcpBaker:94} --
we call it ``fancy'' backtracking -- is also implemented and compared
to the other search strategies. It differs from the original dynamic
backtracking only in the unlabelling procedure: together with the most
recent assignment involved in a detected dead end, all assignments
that are directly or indirectly determined by this assignment are
deleted, too.

\begin{example}
  Let us assume that the most recent assignment involved in a dead end
  is that to the variable~$V_7$.  Further, let us assume that its
  justification is in the elimination explanation of the value~0 of
  the labelled variable~$V_5$.  Then, the assignment of the value 1 to
  the variable~$V_5$ is determined by the assignment to the
  variable~$V_7$. Furthermore, any elimination explanation, e.g.  that
  of the value~1 to the variable~$V_9$, containing the justification
  of the assignment to the variable~$V_5$ is indirectly determined by
  the assignment to the variable~$V_7$. \qed
\end{example}    

Figure~\ref{fbtUnlabel} shows our implementation of this proposed
extension of dynamic backtracking.  The additional loop
(Figure~\ref{fbtUnlabel}, lines 18--33) calculates the set of all
justifications of the assignments that must be deleted containing
at least the justification of the most recent assignment involved
in the detected dead end (Figure~\ref{fbtUnlabel}, line 16). Then, 
all these assignments are deleted (Figure~\ref{fbtUnlabel}, line 34). 
Additionally, the elimination explanations of all variables
are updated, i.e. all defined elimination explanations disjoint to 
the justifications of the deleted assignments are kept because they are
still valid (Figure~\ref{fbtUnlabel}, lines 23--25 and 45--47).

\begin{figure}
\begin{quote}
\txt{int fbtUnlabel(int i)} \BEGIN \\
(01)\tab \txt{Variable var = unlabelledVars.removeLast();}\\
(02)\tab \IF \txt{(var.elimExpl[0] == $\emptyset$ \&\& var.elimExpl[1] == $\emptyset$)} \BEGIN\\
(03)\tab\tab \RETURN \txt{0;}\\
(04)\tab \END\\
(05)\tab \txt{Variable bt;}\\
(06)\tab \txt{int h = $\max(\txt{var.elimExpl[0]} \cup \txt{var.elimExpl[1]})$;}\\
(07)\tab \FOR \txt{(int j=labelledVars.size()-1; j >= 0; j--)} \BEGIN\\
(08)\tab\tab \txt{bt = labelledVars.get(j);}\\
(09)\tab\tab \IF \txt{(bt.num == h)} \BEGIN\\
(10)\tab\tab\tab \txt{labelledVars.remove(j);}\\
(11)\tab\tab\tab \txt{break;}\\
(12)\tab\tab \END\\
(13)\tab \txt{bt.elimExpl[bt.value()]}\\
(14)\tab\tab \txt{= (var.elimExpl[0] $\cup$ var.elimExpl[1])$\setminus\{h\}$;}\\
(15)\tab \txt{unlabelledVars.add(bt);}\\
(16)\tab \txt{IntegerSet label = $\{h\}$;}\\
(17)\tab \txt{boolean isChanged = true;}\\
(18)\tab \WHILE \txt{(isChanged)} \BEGIN \\
(19)\tab\tab \txt{isChanged = false;}\\
(20)\tab\tab \FOR \txt{(j=0; j < labelledVars.size(); j++)} \BEGIN\\
(21)\tab\tab\tab \txt{var = labelledVars.get(j);}\\
(22)\tab\tab\tab \FOR \txt{(int k=0; k <= 1, k++)} \BEGIN \\
(23)\tab\tab\tab\tab \IF \txt{(var.elimExpl[k] != null}\\
(24)\tab\tab\tab\tab\tab\tab \txt{\&\& label $\cap$ var.elimExpl[k] != $\emptyset$ } \BEGIN\\ 
(25)\tab\tab\tab\tab\tab \txt{ var.elimExpl[k] = null;}\\ 
(26)\tab\tab\tab\tab\tab \txt{label = label $\cup~\{\txt{var.num}\}$;}\\
(27)\tab\tab\tab\tab\tab \txt{labelledVars.remove(j);}\\
(28)\tab\tab\tab\tab\tab \txt{unlabelledVars.add(var);}\\
(29)\tab\tab\tab\tab\tab \txt{isChanged = true;}\\
(30)\tab\tab\tab\tab \END\\
(31)\tab\tab\tab \END\\
(32)\tab\tab \END\\
(33)\tab \END\\
(34)\tab\txt{\underline{cs.delete(label)};}\\
(35)\tab \FOR \txt{(j=labelledVars.size()-1; j >= 0; j--)} \BEGIN\\
(36)\tab\tab \txt{var = labelledVars.get(j);}\\
(37)\tab\tab\tab\tab \IF \txt{(!var.isBound())} \BEGIN\\
(38)\tab\tab\tab\tab\tab \txt{labelledVars.remove(j);}\\ 
(39)\tab\tab\tab \txt{unlabelledVars.add(var);}\\ 
(40)\tab\tab \END\\
(41)\tab \END\\
(42)\tab \FOR \txt{(j=0; j < unlabelledVars.size(); j++)} \BEGIN\\
(43)\tab\tab \txt{var = unlabelledVars.get(j);}\\
(44)\tab\tab \FOR \txt{(int k=0; k <= 1, k++)} \BEGIN \\
(45)\tab\tab\tab \IF \txt{(var.elimExpl[k] != null}\\
(46)\tab\tab\tab\tab\tab \txt{\&\& var.elimExpl[k] $\cap$ label != $\emptyset$)} \BEGIN\\ 
(47)\tab\tab\tab\tab \txt{ var.elimExpl[k] = null;}\\ 
(48)\tab\tab\tab \END\\
(49)\tab\tab \END\\
(50)\tab \RETURN \txt{labelledVars.size()+1;}\\
\END\\
\end{quote}
\caption{The alternative unlabelling method of a variant of dynamic backtracking deleting 
  additional variable assignments also keeping formerly detected and
  still valid elimination explanations}\label{fbtUnlabel}
\end{figure}

\section{Performance Comparison}\label{experiments}

We have compared the different search procedures presented in
Section~\ref{search} together with a SICStus Prolog implementation of
chronological backtracking based on the SICStus Prolog compilation of
the CHR handler for Boolean constraints presented in
Section~\ref{boolSolver}. We applied these procedures, then, to all
AIM instances with 50 variables.\footnote{Larger instances tended to
  take too much time (some over 24 hours).} For each instance, the
required backtracking or backjumping steps together with their elapsed
runtime are listed in Tables~\ref{table1} and~\ref{table2}.  The
runtime was measured on a Pentium~IV~PC with 2.8~GHz running Windows~XP 
Professional, Java~1.4.0 from Sun\footnote{see {\tt
    http://java.sun.org/}} and SICStus Prolog\footnote{see {\tt
    http://www.sics.se/sicstus/}}~3.11.0.  In Table~\ref{table1}, CBJ means
conflict-directed backjumping introduced by~\cite{dcpProsser:93} as
presented in Section~\ref{secCBJ}, DBT means dynamic backtracking
introduced in~\cite{dcpGinsberg:93} while FBT is its ``fancy'' variant
introduced in~\cite{dcpBaker:94}, both presented in
Section~\ref{secDBT}.  In Table~\ref{table2}, both chronological
backtracking procedures -- the one presented in Section~\ref{secCBT}
and the other implemented in SICStus Prolog -- obviously require the
same number of search steps; they differ only in their runtime.
 
\begin{table}
\caption{Comparison of different search strategies on the AIM instances 
with 50 variables (Part 1)}\label{table1}
\begin{minipage}{\textwidth}
\begin{tabular}{lrrrrrr}
  \hline \hline  
 &  \multicolumn{6}{c}{Intelligent Backtracking} \\
 AIM instance   & \multicolumn{2}{c}{CBJ\footnote{Conflict-directed 
Backjumping as presented in Section~\ref{secCBJ}}} 
                & \multicolumn{2}{c}{DBT\footnote{original version of 
Dynamic Backtracking~\cite{dcpGinsberg:93}}} 
                & \multicolumn{2}{c}{FBT\footnote{extended version of 
Dynamic Backtracking~\cite{dcpBaker:94}}} \\
\\                
& steps\footnote{backjumping/backtracking steps required to find the
solution or detect the inconsistency} 
& msecs.\footnote{elapsed runtime on a Pentium~IV~PC with 2.8~GHz running
Windows~XP Professional} & steps & msecs. & steps & msecs. \\
\hline
aim-50-1\underline{~}6-yes1-1.cnf&2807&2859&60469&55250&74883&63062\\
aim-50-1\underline{~}6-no-1.cnf&1070&1219&1899&1266&2343&1609\\
aim-50-1\underline{~}6-yes1-2.cnf&772&1031&308192&180172&335314&230983\\
aim-50-1\underline{~}6-no-2.cnf&116&281&456&469&727&687\\
aim-50-1\underline{~}6-yes1-3.cnf&3&63&3&47&3&63\\
aim-50-1\underline{~}6-no-3.cnf&1282&1281&186652&95250&2103&1375\\
aim-50-1\underline{~}6-yes1-4.cnf&287&484&1219&1219&1458&1578\\
aim-50-1\underline{~}6-no-4.cnf&62&203&510&531&87&218\\
aim-50-2\underline{~}0-yes1-1.cnf&200&406&3124&2734&6292&5094\\
aim-50-2\underline{~}0-no-1.cnf&23899&27843&28552&26421&96102&104983\\
aim-50-2\underline{~}0-yes1-2.cnf&98&266&83&218&124&296\\
aim-50-2\underline{~}0-no-2.cnf&409&765&7083&8860&2082&2875\\
aim-50-2\underline{~}0-yes1-3.cnf&130&312&772&1203&7516&10407\\
aim-50-2\underline{~}0-no-3.cnf&28414&30031&287093&28779&257828&271545\\
aim-50-2\underline{~}0-yes1-4.cnf&88&281&135&297&124&297\\
aim-50-2\underline{~}0-no-4.cnf&132&375&3367&3734&4800&6391\\
aim-50-3\underline{~}4-yes1-1.cnf&187&750&259&907&403&1343\\
aim-50-3\underline{~}4-yes1-2.cnf&2&94&2&78&2&78\\
aim-50-3\underline{~}4-yes1-3.cnf&717&2953&1140&4125&1420&5297\\
aim-50-3\underline{~}4-yes1-4.cnf&147&797&106&563&108&578\\
aim-50-6\underline{~}0-yes1-1.cnf&28&422&28&437&28&437\\
aim-50-6\underline{~}0-yes1-2.cnf&13&250&24&359&24&343\\
aim-50-6\underline{~}0-yes1-3.cnf&20&313&24&313&30&343\\
aim-50-6\underline{~}0-yes1-4.cnf&7&234&7&250&7&235\\
\hline
$\sum$ yes1-instances&5506&11515&375587&248172&427736&320434\\
$\sum$ no-instances&55384&61998&515612&165310&366072&389683\\
\hline\hline
\end{tabular}
\vspace{-2\baselineskip}
\end{minipage}
\end{table}

\begin{table}
\caption{Comparison of different search strategies on the AIM instances 
with 50 variables (Part 2)}\label{table2}
\begin{minipage}{\textwidth}
\begin{tabular}{lrrr}
  \hline \hline  
               & \multicolumn{3}{c}{Chronological Backtracking} \\
  AIM instance & common & \multicolumn{1}{c}{Adaptive CHR in Java} 
                        & \multicolumn{1}{c}{CHR in SICStus Prolog}\\
               & steps\footnote{backtracking steps required to find the 
                 solution or detect the inconsistency}  
               & msecs.\footnote{elapsed runtime on a Pentium~IV~PC with 
                 2.8~GHz running Windows~XP Professional}                      & msecs. \\
\hline
aim-50-1\underline{~}6-yes1-1.cnf&86442&56610&56500\\
aim-50-1\underline{~}6-no-1.cnf&1355146&680759&571907\\
aim-50-1\underline{~}6-yes1-2.cnf&402870&147620&129030\\
aim-50-1\underline{~}6-no-2.cnf&309298&151918&155876\\
aim-50-1\underline{~}6-yes1-3.cnf&3&47&10\\
aim-50-1\underline{~}6-no-3.cnf&6213098&2009531&1461955\\
aim-50-1\underline{~}6-yes1-4.cnf&22684&13484&14062\\
aim-50-1\underline{~}6-no-4.cnf&1152796&432173&316532\\
aim-50-2\underline{~}0-yes1-1.cnf&8936&7781&9640\\
aim-50-2\underline{~}0-no-1.cnf&536726&305183&262791\\
aim-50-2\underline{~}0-yes1-2.cnf&305&453&296\\
aim-50-2\underline{~}0-no-2.cnf&59470&55093&65421\\
aim-50-2\underline{~}0-yes1-3.cnf&21549&20827&26406\\
aim-50-2\underline{~}0-no-3.cnf&127034&106859&105455\\
aim-50-2\underline{~}0-yes1-4.cnf&217&469&312\\
aim-50-2\underline{~}0-no-4.cnf&45542&39657&37081\\
aim-50-3\underline{~}4-yes1-1.cnf&352&1156&1497\\
aim-50-3\underline{~}4-yes1-2.cnf&2&78&30\\
aim-50-3\underline{~}4-yes1-3.cnf&916&3547&5158\\
aim-50-3\underline{~}4-yes1-4.cnf&281&1156&1561\\
aim-50-6\underline{~}0-yes1-1.cnf&28&438&672\\
aim-50-6\underline{~}0-yes1-2.cnf&15&265&329\\
aim-50-6\underline{~}0-yes1-3.cnf&47&469&782\\
aim-50-6\underline{~}0-yes1-4.cnf&7&250&236\\
\hline
$\sum$ yes1-instances&544654&254650&427464\\
$\sum$ no-instances&9799110&3781173&2977018\\
\hline\hline
\end{tabular}
\vspace{-2\baselineskip}
\end{minipage}
\end{table}

\begin{figure}
\epsfig{file=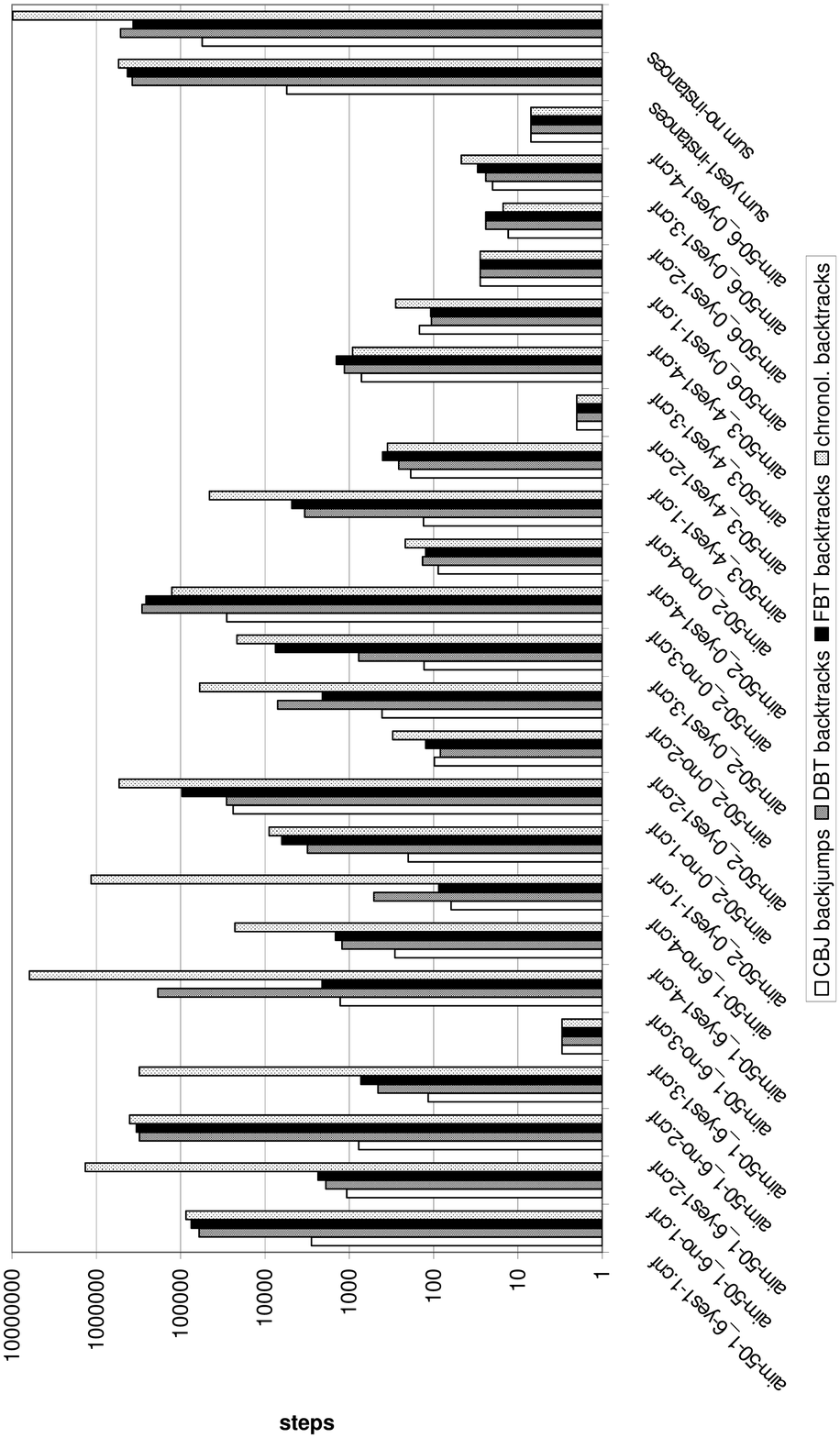, angle=0, clip=, height=1.0\textheight}
\caption{Comparison of different search strategies on the AIM instances with 50 variables in terms of their 
         number of backjumping/backtracking steps}\label{diag1}
\end{figure}

\begin{figure}
\epsfig{file=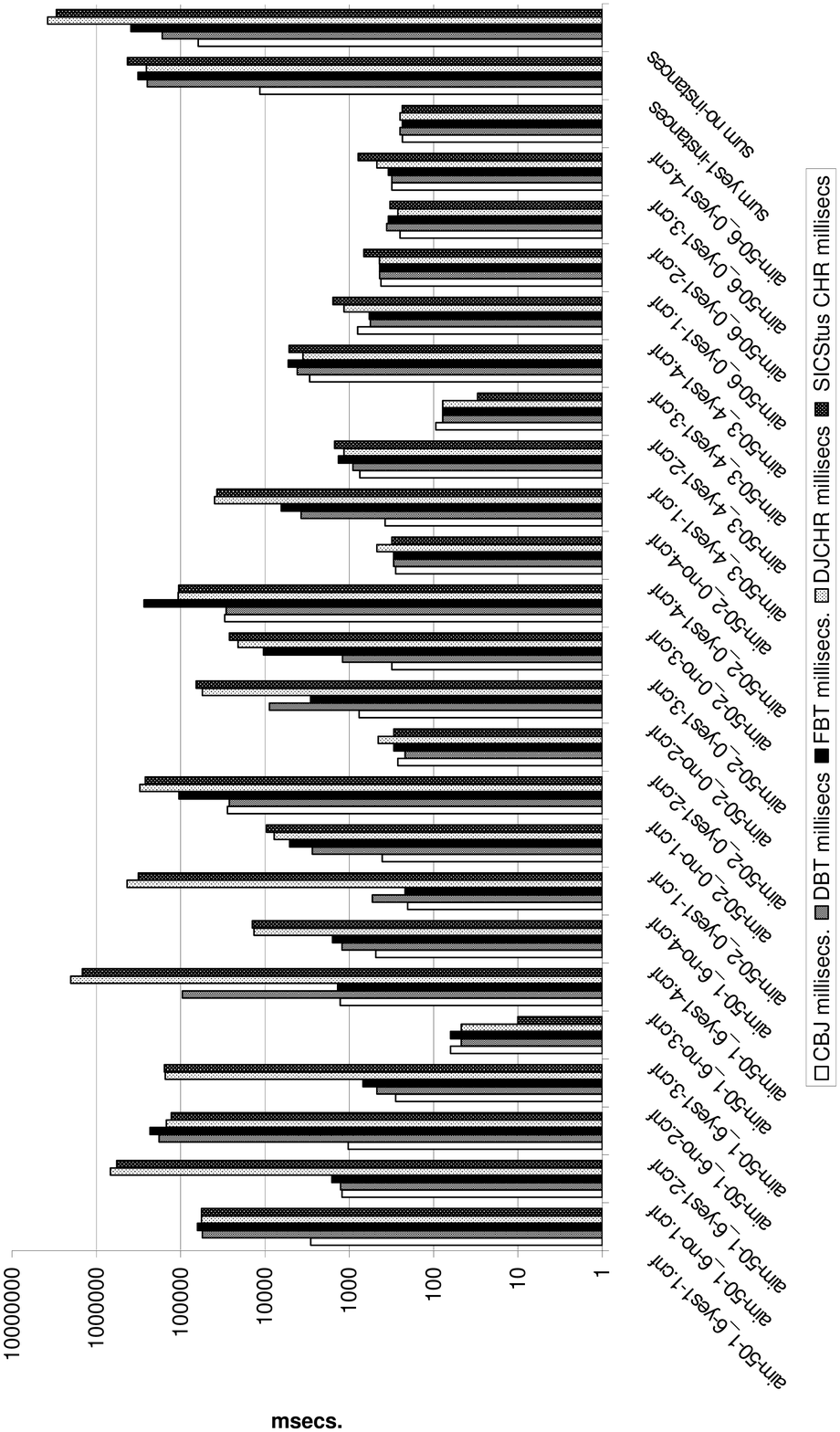, angle=0, clip=, height=1.0\textheight}
\caption{Comparison of different search strategies on the AIM instances with 50 variables in terms of their runtime}\label{diag2}
\end{figure}

Based on these results, we compared the number of search steps and
runtime in graph form.  Figure~\ref{diag1} shows the ``qualitative''
comparison, and Figure~\ref{diag2} the ``quantitative'' comparison. In
both figures, the last two groups show the summations of the steps/the
elapsed runtime for the AIM instances with exactly one solution
(yes-instances) and for the inconsistent instances (no-instances). The
summations show that conflict-directed backjumping performs very well,
confirming the results in~\cite{dcpProsser:93}: in terms of the number
of search steps, conflict-directed backjumping (CBJ) requires on
average two orders of magnitude less than chronological backtracking
for all instances and is on average more than one order of magnitude
faster than all other search procedures, even faster than the SICStus
Prolog implementation. Furthermore, the performance of the Java and
SICStus Prolog implementations of chronological backtracking are
comparable. Looking at the required number of search steps, we find
that in a few cases dynamic backtracking (DBT) requires marginally
more steps than chronological backtracking, which is at odds with the
statements made in~\cite{dcpBaker:94}.  Surprisingly, its extended
variant (FBT) proposed in~\cite{dcpBaker:94} often requires more
search steps than the original version of dynamic backtracking.

\section{Discussion}\label{discussion}

The conflict-directed backjumping algorithms presented
in~\cite{dcpProsser:93,dcpProsser:95} compute the conflict sets either
from total assignments with respect to the violated constraints or by
using forward checking or maintenance of arc consistency,
respectively.  The calculation of the elimination explanations
in~\cite{dcpBaker:94,dcpGinsberg:93,dcpJussien:etal:00} during
practical experiments was rather similar. Thus, we assume that
inconsistencies are detected rather late, after a lot of superfluous,
unsuccessful assignments, i.e. search steps. In our approach, the
underlying Boolean constraint solver performs local, but also some
``global'' constraint propagation (cf. Example~\ref{mac-expl}). In
general, the search spaces are more restricted. Thus, we expect
inconsistencies to be detected earlier in the search tree, resulting
in more general conflict sets or elimination explanations and also
earlier detection of dead ends.

\begin{example}
  Considering the Boolean constraint solver presented in
  Section~\ref{boolSolver} and the constraints
\[
  \mbox{\tt neg(X,Y), or(X,Y,Z), neg(Z,U)}\enspace,
\]
the application of one of the CHR on the first and second constraint
will add the syntactical equation {\tt Z=1}. Further addition of the
assignment, i.e. equation {\tt U=1}, will result in an inconsistency,
i.e. {\tt false}, by applying one of the CHR to the actualised third
constraint, i.e. {\tt neg(1,1)}. Thus, in conflict-directed
backjumping and in dynamic backtracking, the assignment {\tt U=1} is
excluded during any further search: the conflict set and
elimination explanation of value~1 of the variable {\tt U} will be
the empty set. Or, if we consider the constraints 
\[
\mbox{\tt or(X,Y,Z), neg(Z,1)}\enspace,
\]
the assignment {\tt X=1} triggers a CHR that derives the equation {\tt
  Z=1}, resulting in the constraint {\tt neg(1,1)} and eventually in
{\tt false}. Now, the assignment {\tt X=1} is excluded during any
further search, too.  \qed
\end{example}

We assume that this kind of ``consistency maintenance'', i.e.
constraint propagation, is - at least partially - responsible for the
absence of the bad performance of dynamic backtracking when applied to
3-SAT problems, as reported in~\cite{dcpBaker:94}.

\section{Conclusion}\label{conclusion}

During our review of the adaptive CHR system, we have emphasised the
potential of this system for explanation-based constraint
programming~\cite{dcpJussien:01}. One possibility here is the use of
explanations in building explanation-guided search algorithms.  More
specifically, we have demonstrated the simplicity of implementing
sophisticated ``intelligent'' search strategies in conjunction with a
CHR-based constraint solver within this system. In this context,
\begin{itemize}
  \item ``simplicity'' means that the implementations are quite 
        straightforward, 
        using the interface to the underlying adaptive constraint solver 
        in an obvious manner
  \item ``sophisticated'' means that early inconsistency detection
        accomplished by constraint propagation within the underlying 
        solver obviously reduces the number of search steps
\end{itemize} 
Conflict-directed backjumping and dynamic backtracking based on CHR
thus gain a kind of ``consistency maintenance'' and the poor
performance of dynamic backtracking reported in~\cite{dcpBaker:94}
does not occur.
        
An empirical comparison of the implemented search procedures on the
AIM instances showed that the addition of ``intelligence'' to the
search process may reduce the number of search steps dramatically.
Even the rather simple conflict-directed backjumping strategy
outperforms on average all the other strategies tested. Furthermore, we
have shown that the runtime of the Java implementations of the
intelligent search strategies is in most cases better than the
implementations of chronological backtracking, even better than the
implementation in SICStus Prolog.

One of the main conclusions in a recent paper on building
state-of-the-art SAT solvers~\cite{satLynce:MarquesSilva:02}
``\emph{\ldots is that applying non-chronological backtracking is most
  often crucial in solving real-word instances of SAT.}'' -- Our
implemented ``intelligent'' search procedures belong to this set of
non-chronological backtracking solvers. A further development of the
presented techniques~\cite{cpMueller:04} shows that their performance
is comparable to those of these state-of-the-art SAT solvers.

\section{Future Work}

Future work will focus on fast implementation techniques, like
counter-based or lazy implementations, randomised and heuristic
variable selection, and assignment strategies that are commonly used in
other state-of-the-art SAT solvers~\cite{satLynce:MarquesSilva:02} as
well as the implementation of \emph{partial order dynamic
  backtracking}~\cite{dcpGinsberg:McAllester:1994} or its
generalisation~\cite{dcpBliek:98}. Furthermore, all these extensions
and the presented search strategies will be compared with some local
search algorithms that might also be implemented on the basis of our
adaptive CHR system~\cite{chrWolf:01a}.  Further future research
topics are the implementation of the generalisations proposed during
our presentation of the search strategies and their application to and
comparison ´with other finite-domain constraint satisfaction problems
like job-shop scheduling.

Conflict-directed backjumping performs well for \emph{Quantified
  Boolean Logic Satisfiability}~\cite{satGiunchiglia:01}. The
algorithm presented here might therefore be adapted and successfully
applied to this problem class, which is strongly related to SAT.

\bibliography{atms,ccp,chr,classics,clp,cp,dedu,dyncp,hclp,java,lp,unify,sat}


\end{document}